\documentclass[journal]{IEEEtran}

\ifCLASSINFOpdf

\else

\fi
\usepackage{amsfonts}

\usepackage{adjustbox}
\usepackage{algorithmicx}
\usepackage{algpseudocode}
\usepackage[ruled]{algorithm}
\usepackage{graphicx}
\usepackage{caption}
\usepackage{graphicx,times,amsmath} 
\usepackage{caption}
\usepackage{subcaption}
\usepackage[rightcaption]{sidecap}
\usepackage{caption}
\usepackage{multirow}
\usepackage{hhline}
\usepackage{amsmath}
\usepackage{epstopdf}
\usepackage{tabularx}
\usepackage[font=footnotesize]{caption}
\usepackage{cite}
\graphicspath{ {images/} }

\usepackage{fancyhdr}
\fancypagestyle{pageStyleOne}{%
\normalfont\footnotesize
    \fancyhf{}
\fancyhead[L]{\fontsize{8}{8} \selectfont This paper is submitted for peer-review.}
}

\makeatletter

\newcommand{\norm}[1]{\left\lVert#1\right\rVert}

\begin{document}

\title{Interpretation of Mammogram and Chest Radiograph Reports Using Deep Neural Networks}

\author{Hojjat Salehinejad,
        Shahrokh Valaee,~\IEEEmembership{Senior Member,~IEEE},
        Aren Mnatzakanian,      
        Tim Dowdell,\\
        Joseph Barfett, and
        Errol Colak \vspace{-8mm}
\thanks{H. Salehinejad is with the Edward S. Rogers Sr. Department of Electrical \& Computer Engineering, University of Toronto, Toronto, Canada, and Department of Medical Imaging, St. Michael's Hospital, University of Toronto, Toronto, Canada, e-mail: salehinejadh@smh.ca.}
\thanks{S. Valaee is with the Edward S. Rogers Sr. Department of Electrical \& Computer Engineering, University of Toronto, Toronto, Canada, e-mail: valaee@ece.utoronto.ca.}
\thanks{A. Mnatzakanian, T. Dowdell, J. Barfett, and E. Colak are with the Department of Medical Imaging, St. Michael's Hospital, University of Toronto, Toronto, Canada, e-mail: \{mnatzakaniana,dowdellt,barfettj,colake\}@smh.ca.}
\thanks{}}




\maketitle
\thispagestyle{pageStyleOne}

\begin{abstract}
Radiology reports are an important means of communication between radiologists and other physicians. These reports express a radiologist's interpretation of a medical imaging examination and are critical in establishing a diagnosis and formulating a treatment plan. In this paper, we propose a Bi-directional convolutional neural network (Bi-CNN) model for the interpretation and classification of mammograms based on breast density and chest radiographic radiology reports based on the basis of chest pathology. The proposed approach is a part of an auditing system that evaluates radiology reports to decrease incorrect diagnoses. Our study revealed that the proposed Bi-CNN outperforms convolutional neural network, random forest and support vector machine methods.


\end{abstract}


%
\IEEEpeerreviewmaketitle

 \vspace{-5mm}
\section{Introduction}
\label{sec:introduction}

Radiology reports are an important means of communication between radiologists and other physicians \cite{naik2001radiology}. These reports express a radiologist's interpretation of a medical imaging examination and are critical in establishing a diagnosis and formulating a treatment plan. 

Radiology is among the medical specialties with the highest rate of malpractice claims \cite{pinto2010}. These claims can arise from a failure to communicate important findings, a failure of perception, lack of knowledge, and misjudgment. A failure to detect an abnormality on a medical imaging examination can lead to significant medical consequences for patients such as a delayed diagnosis. With a delayed or incorrect diagnosis, patients can present later with worsened symptoms and more advanced disease that may require more aggressive treatment or may be untreatable.

In this paper, we propose an auditing system for radiologists that has two main components: a natural language processing (NLP) model to process and interpret a radiology report and a machine vision model that interprets the medical imaging examination. This auditing system reviews the radiologist's report and compares it with the interpretation of a machine vision model. The proposed system would notify the radiologist if there is a discrepancy between the two interpretations. Many investigators have aimed to develop machine vision models for the interpretation of medical imaging examinations, such as chest radiographs \cite{cicero2017training}, mammograms \cite{wang2017detecting}, and head computerized tomography (CT) scans \cite{havaei2017brain}. However, fewer attempts have been made for the NLP of radiology reports \cite{aronow1999ad, wilcox2000automated, shin2017classification}. The focus of this paper is on the design and performance evaluation of the NLP component.

We propose a bi-directional convolutional neural network (Bi-CNN) for the NLP of radiology reports. The Bi-CNN will be trained independently for two distinct report data types: breast mammograms and chest radiographs. In particular, this model will classify the degree of breast density and the type of chest pathology based on the radiology report content. Our proposed NLP model differs from a keyword search algorithm. It is capable of interpreting and classifying a radiology report. The proposed Bi-CNN has two independent input channels, where the order of non-padded input to one channel is the reverse of the non-padded input to the other channel. Performance of the proposed model is compared with a single channel convolutional neural network (CNN), random forest (RF), and support vector machine (SVM) models and a comparative study is conducted.

\begin{figure*}[!htbp]
\centering
\captionsetup{font=footnotesize}
                \includegraphics[width=0.7\textwidth]{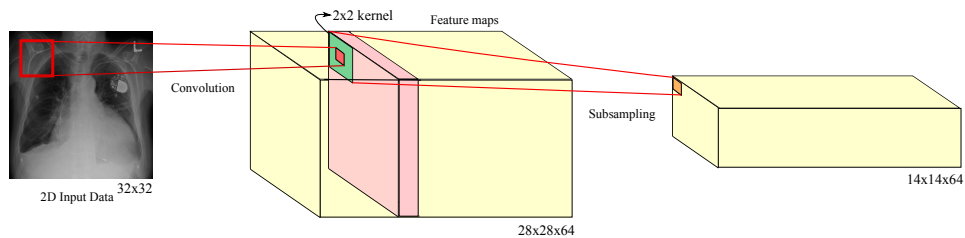}
\caption{Convolution and sub-sampling in convolutional neural networks.}
\label{fig:simplecnn}
 \vspace{-5mm}

\end{figure*}

\section{Related Work}
\label{sec:relatedworks}
The NLP models used for the classification and interpretation of radiology reports can be categorized into rule-based and machine learning models. Rule-based systems are usually built upon a series of ``if-then'' rules, which require an exhaustive search through documents. Such rules are typically designed by human experts. By contrast machine learning models do not require explicit rules. Instead, they try to train themselves iteratively and extract features from data. 

\subsection{Rule-based Approaches}
A radiology report mining system generally has three main components. These include a medical finding extractor, a report and image retriever, and a text-assisted image feature
extractor \cite{gong2008text}. For example, a set of hand-crafted semantic rules are presented in \cite{gong2008text} for mining brain CT radiology reports. A natural language parser for medical reports is proposed in \cite{taira2007field} that uses four-gram and higher order systems to assign a stability metric of a reference word within
a given sentence. This model is inspired from Field theory and uses a dependency tree that represents
the global minimum energy state of the system of words for
a given sentence in radiology reports \cite{taira2007field}. A classifier using extracted keywords from clinical chest radiograph reports is developed in \cite{cooper1998using}. In \cite{chapman2003creating}, a method is developed to identify patients with mediastinal findings related to inhalational anthrax. These methods are applied on different case studies with different sets of rules. Such methods are not flexible enough to be applied to case studies for which they were not designed, and hence are not transferable, and their performance comparison requires specific implementation and fine tuning of rules.

\subsection{Machine Learning Approaches}
A NLP study for a large database of radiology reports was conducted in \cite{dang2009use} based on two classes: positive and negative radiology findings. These classes have different patient attributes such as age groups, gender, and clinical indications \cite{dang2009use}. A collection of 99 musculoskeletal radiology examinations were studied using a machine learning model versus a naive Bayes classification algorithm, followed by a SVM to detect limb fractures from radiology reports \cite{zuccon2013automatic}. A CNN was used in \cite{shin2015interleaved} as a supervised tool to map from images
to label spaces. This system has an interactive component between supervised and unsupervised machine learning models to extract
and mine the semantic interactions of radiology images and
reports. Recurrent neural networks (RNN) along with the CNN have been used to interpret chest x-ray reports for a limited number of classes \cite{shin2016learning}. This CNN model learns from the annotated sequences produced by the RNN model. The RNN model with long short term memory (LSTM) is capable of learning long term dependencies between the annotations \cite{salehinejad2016learning}. The model proposed in \cite{lai2015recurrent} uses a recurrent design to extract features from word representations. Compared with window-based neural networks, which uses a rolling window to extract features, this model generates less noise and is more stable. The max-pooling method distinguishes the words that have a major impact in the report.  
Machine learning models have also been developed for understanding medical reports in other languages, Korean being one example, using tools such as Naive Bayes classifiers,
maximum entropy, and SVMs \cite{oh2011extracting}.

\subsection{Learning Vector Representations of Words}
Word2vec is a well-known model used for learning vector space representations of words. It produces a vector space from a large corpus of text with reduced dimensionality. This model assigns a word vector to each unique word in the corpus. Vectors with closer contexts are positioned in closer proximity in this space \cite{mikolov2013distributed}. 
GloVe approach provides a global vector space representation of words. It uses global matrix factorization and local context windowing methods to construct a global log-bilinear regression model \cite{pennington2014glove}. The word-word co-occurrence matrix of the words is a sparse matrix by nature. However, GloVe only uses the non-zero elements for training and does not consider individual context windows in a large corpus \cite{pennington2014glove}. CNNs have been shown to work well on $n$-gram representations of data \cite{majumder2017deep}. For example, a convolution layer can perform feature extraction from various $n$ values of a $n$-gram model and perform personality detection from documents \cite{majumder2017deep}. Dynamic k-Max pooling can operate as a global operation over linear sequences. Such dynamic CNNs can perform semantic modelling of sentences \cite{kalchbrenner2014convolutional}. The model receives variable dimension size input sentences and induces a feature graph over the sentence that is capable of explicitly capturing short and long-range relations \cite{kalchbrenner2014convolutional}.

CNN models are also utilized for learning representations, opinion sentiment understanding, and analysis of products. For example, a CNN with a single output softmax layer can classify a vector representation of text with high accuracy \cite{kim2014convolutional}. The design of output layers in CNNs varies depending on the application of the model. A collaboration of RNNs and CNNs for feature extraction of sentences has been examined, where a CNN learns features from an input sentence and then a gated RNN model discourses the information \cite{ren2017neural}. In such system, a bi-directional long short term memory (Bi-LSTM) RNNs can sequentially learn words from question-answer sentences. The trained network can select an answer sentence for a question and present the corresponding likelihood for correctness \cite{Wang2015ALS}. 

\begin{figure*}[!htbp]
\centering
\captionsetup{font=footnotesize}
                \includegraphics[width=0.7\textwidth]{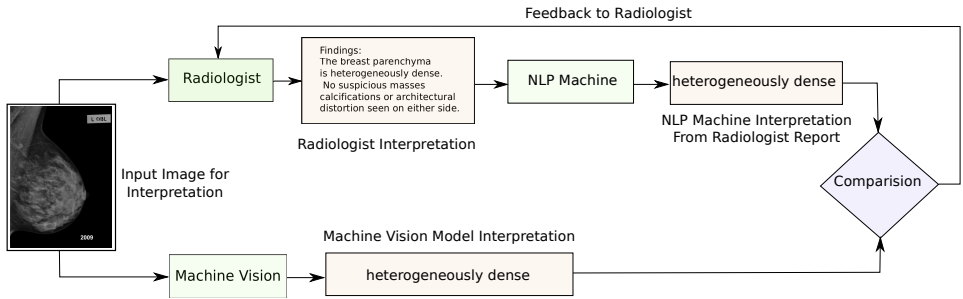}
\caption{Block-diagram of the proposed auditing system.}
\label{fig:audiotory}
 \vspace{-5mm}

\end{figure*}

\section{A Brief Review on Convolutional Neural Networks}
\label{sec:cnn}
As NLP for radiology reports is an interdisciplinary research effort, a brief review on CNN is provided in this section. A CNN is a machine learning model inspired by the visual cortex of cats \cite{hubel1968receptive}. A CNN has at least one layer of convolution and sub-sampling, in which the number of layers can increase in depth, generally followed by a fully connected multi-layer perceptron (MLP) network. A consecutive arrangement of convolutional layers followed by sub-sampling build a pyramid-shape model, where the number of feature maps increases as the spatial resolution decreases. Instead of hand-designing feature extractors, the convolutional layers extract features from raw data and the MLP network classifies the features \cite{lecun1995convolutional}. 
A CNN typically has three main pieces which are local receptive fields, shared weights, and spatial and/or temporal sub-sampling.

\subsection{Local Receptive Fields}
Local receptive fields are made from artificial neurons, which observe and extract features from data such as edges in images. Let us consider an image as input to a CNN such as in Figure~\ref{fig:simplecnn}. A rectangular kernel with size $M\times N$ scans the input matrix at every single element (i.e., a pixel) and performs convolution such as 
\begin{equation}
h_{u,v} = \sigma(\sum_{m=0}^{M-1}\sum_{n=0}^{N-1}(I_{u+m,v+n}\cdot w_{m,n})+b_{u,v})
\end{equation}
where the output is the state of the neurons at element $I_{u,v}$, $\sigma(\cdot)$ is a non-linear function, $\textbf{W}$ is the shared weights matrix, and $\textbf{b}$ is the bias vector.

\subsection{Shared Weights}
The term ``shared weights" refers to the weight matrix $\textbf{W}$ which appears repeatedly in the convolution operation for every element of $I$. Weight sharing reduces the number of free parameters of the model and hence improves generalization of the model  \cite{lecun1995convolutional}.

For a general machine vision task, such as classification of natural images, the convolution kernel uses the correlation among spatial or temporal elements of the image to extract local features. Since the image is stationary (i.e., statistics of one part of the image are similar to any other part), the learned shared weights using one sub-region of the image can also be used at another sub-region to extract different features.

\subsection{Sub-sampling}
Passing the whole bag of extracted features after the convolution operation to a classifier is computationally expensive \cite{boureau2010theoretical}. The feature pooling task (i.e. sub-sampling) generalizes the network by reducing the resolution of the dimensionality of intermediate representations (i.e. feature maps) as well as the sensitivity of the output to shifts and distortions \cite{lecun1995convolutional}. The two most popular subsampling methods are mean-pooling and max-pooling. By dividing the feature map into a number of non-overlapping rectangle-shape sub-regions, the mean-pooling computes the average of features sitting inside a sub-region as the pooled feature of that sub-region. Max-pooling performs the same task as mean-pooling but with a maximum operator. A study on these two operations is provided in \cite{boureau2010theoretical}.

\begin{figure*}[!htbp]
\centering
\captionsetup{font=small}
                \includegraphics[width=0.8\textwidth]{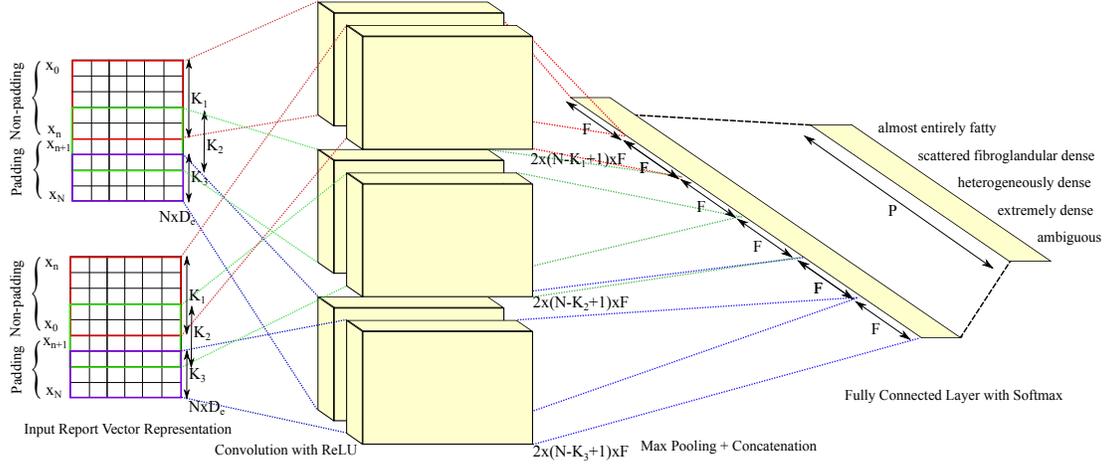}
\caption{Proposed bi-directional convolutional neural network for the interpretation and classification of mammograms based on breast density. For the chest reports case, the labels of the output layer change accordingly but the model architecture is identical. }
\label{fig:cnn}
 \vspace{-5mm}
\end{figure*}

\begin{figure}[!htbp]
\centering
\captionsetup{font=small}
                \includegraphics[width=0.3\textwidth]{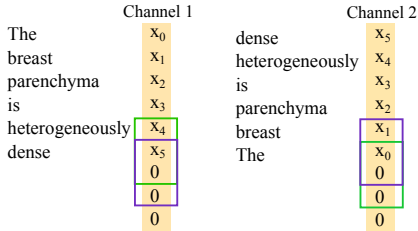}
\caption{The order of feature vectors and padding for the two input channels of bi-directional convolutional neural network (Bi-CNN). Each channel has two kernels (green and purple frames) of size three. The order of feature vectors in Channel 2 is the reverse order of feature vectors in Channel 1.}
\label{fig:reverse}
 \vspace{-5mm}
\end{figure}

\section{The Proposed Method}
\label{sec:methods}
An auditing system for radiology is presented in Figure~\ref{fig:audiotory}. The main components of the system are a machine vision model to interpret the input image and a NLP component to interpret the corresponding radiology report. The system notifies the radiologist if there is a discrepancy between the two components of the system. In this paper, our focus is on the design and evaluation of the NLP component.

\subsection{Preprocessing}
In general, radiology reports include major sections such as ``Indication", ``Findings", ``Impression", and  the name of the reporting radiologist. However, there are variations in report formatting due to ``radiologist style" or institutional requirements. Despite brute-force search approaches, the deep learning models have the advantage of requiring the least amount of data cleaning and preprocessing due to their natural adaptation to the data and non-linear feature extraction ability.

In the preprocessing step, the model extracts the ``Findings" and ``Impression" sections of the report and performs tokenization and string cleaning. Some of the major tasks include converting upper-case characters to lower-case, removing unnecessary punctuation and symbols, and separating the remainder from the attached string. We add every unique word to a vocabulary dictionary. Each vocabulary has a unique associated index which represents it in the sentence vector. For example, given the sentence \textit{``The breasts show scattered fibroglandular tissue."}, the entire vocabulary dictionary would consist of 6 words (breasts, show, scattered, fibroglandular, tissue, the), which is represented as a list of integers such as $[1,...,6]$. A vector to the length of the longest report in terms of words count in the dataset represents the report. Padding fills up the gap for shorter reports. 

In the character embedding step, each integer is mapped to a high dimensional (i.e., $D$) vector with a uniform random distribution. 
In such a high dimensional space, due to the ``curse of dimensionality",  the vectors are considered independent \cite{weber1998quantitative}, \cite{domingos2012few}. Character embedding vector generation for medical vocabularies based on the correlation among the vectors (i.e., similar to Word2Vec) is an interesting topic for further investigation.

\subsection{Bi-CNN Model Architecture}
An input report with $N$ words can be represented as a sequence
$\textbf{X}= [\textbf{x}_{1},... ,\textbf{x}_{N}]$, where each word is a vector $\textbf{x}_{n}\in \mathbb{R}^{D}$, \cite{salehinejad2017convolutional}. As Figure~\ref{fig:cnn} shows, the proposed architecture has two input channels followed by two independent convolution layers. Both channels have an identical design, except the order of non-padded input to one channel is the reverse of the non-padded input to the other channel. In the example in Figure~\ref{fig:reverse}, the dependencies between feature vectors ($\textbf{x}_{3},\textbf{x}_{4},\textbf{x}_{5}$) are visible to the kernels in both channels. However, the dependencies between feature vectors ($\textbf{x}_{1},\textbf{x}_{0},0$) are only visible to the kernels in Channel~2, while these dependencies are not visible to the kernels in Channel~1. Other examples are the feature vectors ($\textbf{x}_{0},0,0$) in Channel~2 and feature vectors ($\textbf{x}_{4},\textbf{x}_{5},0$) and ($\textbf{x}_{5},0,0$) in Channel~1, which are not visible in the corresponding other channel.

Each channel in the proposed model in Figure~\ref{fig:cnn} has three filters with varying window sizes $K\in\{3,4,5\}$ that slide across the input layer \cite{kim2014convolutional}, \cite{salehinejad2017convolutional}. The filters extract features from the input layer to construct feature maps of size $(N-K+1)\times F$, where $F$ is the number of feature maps for each filter. Each feature map $\textbf{h}_{(N-K+1)}$ has its own shared weight $\textbf{W}_{K\times D}$ and bias $\textbf{b}_{ N-K+1}$. The value of a hidden neuron $m$ is

\begin{equation}
h_{m}=\sigma(\sum_{k=1}^{K}\sum_{d=1}^{D} x_{m+k-1,d} \cdot w_{k,d} +b_{m}).
\end{equation}
The window word is $\textbf{x}_{m:m+K-1}$ and $\sigma(\cdot)$ is a rectified linear unit (ReLU) activation function defined as 
\begin{equation}
\sigma(z)=max(0,z)
\end{equation} 
where $z\in\mathbb{R}$. The max pooling \cite{collobert2011natural} extracts the most significant feature from the feature map of a filter as
\begin{equation}
\hat{h}_{l}=max(\{h_{1},..,h_{N-K+1}\}).
\end{equation}
The output of the max-pooling layer contains the max-pooled features $\hat{\textbf{h}}=[\hat{h}_{1},...,\hat{h}_{L}]$ where the length of the feature vectors is $L=C\cdot F$ and $C$ is the number of classes. For example, for the breast density classification with five classes we have $L=5F$ (see Figure~\ref{fig:cnn}).
The features from the input channels are concatenated and passed to a fully connected perceptron network. The output of the softmax layer is the probability distribution over all the labels $P$ (e.g., for mammograms the breast density classes). The value of output unit $p$ is 

\begin{equation}
y_{p} = \phi(\sum_{l=1}^{L} (\hat{h}_{1,l} \cdot r_{1,l})\cdot w_{l,p}^{ho} + b_{p}^{o})
\end{equation}
where $w_{l,p}^{ho}$ is the weight of connection from the hidden unit $l$ in the hidden layer $h$ to the output unit $p$ in the output layer $o$,  $ b_{p}^{o}$ is the bias of the output unit $p$, and $\phi(\textbf{z})$ is the softmax activation function defined as
\begin{equation}
\phi(z_{p})=\frac{e^{z_{p}}}{\sum_{j=1}^{P}e^{z_{j}}} \:\: \textrm{for} \:\: p=[1,...,P].
\end{equation}

\subsection{Training}
Adam optimizer is a first-order gradient-based optimization method with integrated momentum functionality \cite{kingma2014adam}. During training, the momentum helps to diminish the fluctuations in weight changes over consecutive iterations. The drop-out regularization method randomly drops units along with their connections from the 
concatenated layer (i.e. the input layer to classifier) to the output layer using a binary \textit{mask} vector $\textbf{r}_{1\times L}$ with Bernoulli random distribution \cite{srivastava2014dropout}.

\subsection{Evaluation Scheme}
Since we are dealing with discrete categories, we use cross-entropy between a predicted value $y^{(i)}$ from the network and the real label $t^{(i)}$ to measure the loss of networks such as
\begin{equation}
 \mathcal{\hat{L}}(Y, T) = \frac{-1}{R}\sum_{i=1}^{R}t^{(i)}ln(y^{(i)})+(1-t^{(i)})ln(1-y^{(i)})
\end{equation}
where $R$ is the number of reports, $Y$ is the set of network predictions, and $T$ is the set of targets to predict. Adding the weight decay term (i.e., $L_{2}$ regularization) to the loss function helps the network to avoid over-fitting while training such as
\begin{equation}
 \mathcal{L}(Y,  T) =  \mathcal{\hat{L}}(Y, T) + \eta(\norm{\mathbf{W}^{ho}}_{2})
\end{equation}
where $\mathbf{W}^{ho}$ is the weight matrix of connections between the hidden layer $h$ and the output layer $o$ and $\eta$ is the regularization control parameter.

 \begin{table}[!ht]
\renewcommand{\tabcolsep}{1pt}
\captionsetup{font=small}
\caption{Summary of statistics for the mammogram reports dataset (MRD) and chest radiograph reports dataset (CRRD) for the ``findings" and/or ``impression" sections. NR: number of reports; VS: vocabulary size; ANS: average number of sentences per report; ANW: average number of words per report; ASL: average sentence length per report; Mean: sample mean;
Med: sample median; Std: sample standard deviation.}

\begin{center}
\begin{adjustbox}{width=0.48\textwidth}

\begin{tabular}{|c|c|c|c|c|c|c|c|c|c|c|c|}
\hline
\multirow{2}{*}{Dataset} &\multirow{2}{*}{NR} &\multirow{2}{*}{VS} & \multicolumn{3}{c|}{ANS}& \multicolumn{3}{c|}{ANW}& \multicolumn{3}{c|}{ASL} \\ \hhline{~~~---------}
& & & Mean & Median & Std  & Mean & Median & Std & Mean & Median & Std \\ \hline
\hline

MRD	&  4,080 & 1,691 &      3.21 & 3.00 & 1.27          & 30.87 & 27.00 & 10.44      &9.61 & 9.00  & 8.22 \\ \hline	
CRRD            &  1,030    & 772  &      2.01 & 2.00 & 1.07          & 21.46 & 19.00 & 9.21      &8.21 & 8.00  & 7.67 \\ \hline	
\end{tabular}
\end{adjustbox}
\label{T:stats}
\end{center}
 \vspace{-2mm}

\end{table}

\begin{figure}[!htp]
\centering
\captionsetup{font=small}
        \begin{subfigure}[t]{0.120\textwidth}
        \centering
                \includegraphics[width=0.9\textwidth]{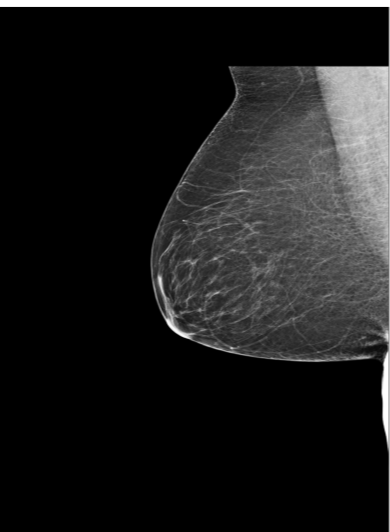}
                \caption{}
                \label{fig:loss}
        \end{subfigure}%
        \begin{subfigure}[t]{0.120\textwidth}
        \centering
                \includegraphics[width=0.9\textwidth]{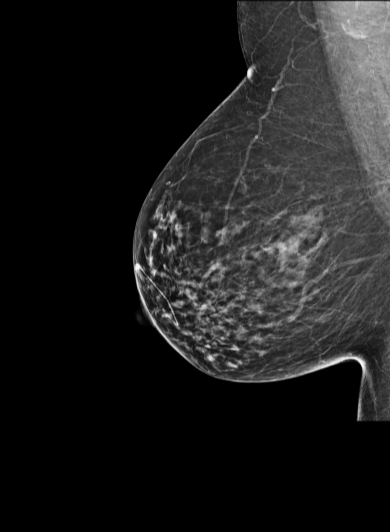}
                \caption{}
                \label{fig:acuracy}
        \end{subfigure}%
        \begin{subfigure}[t]{0.120\textwidth}
        \centering
                \includegraphics[width=0.9\textwidth]{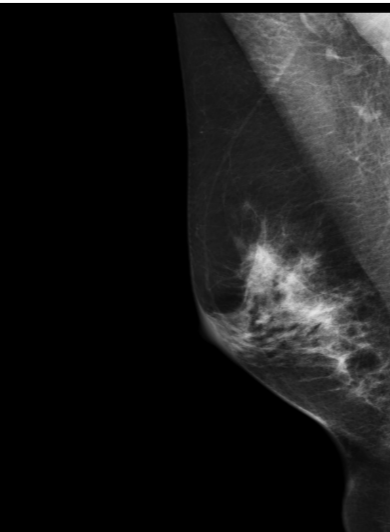}
                \caption{}
                \label{fig:loss}
        \end{subfigure}%
        \begin{subfigure}[t]{0.120\textwidth}
        \centering
                \includegraphics[width=0.9\textwidth,scale=0.2]{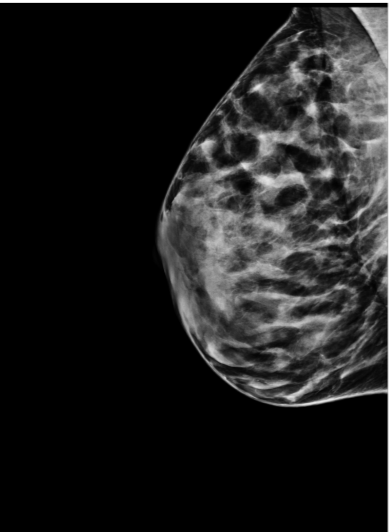}
                \caption{}
                \label{fig:acuracy}
        \end{subfigure}%
        \caption{Four classes of breast density. a) Almost entirely fatty; b)~Scattered areas of fibroglandular density; c) Heterogeneously dense; d) Extremely dense.}
        \label{fig:breast_images} 
\end{figure}
\begin{figure}[!htp]
\centering
\captionsetup{font=small}
        \begin{subfigure}[t]{0.12\textwidth}
        \centering
                \includegraphics[width=0.9\textwidth]{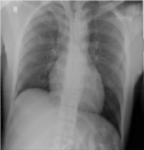}
                \caption{}
                \label{fig:loss}
        \end{subfigure}%
        \begin{subfigure}[t]{0.12\textwidth}
        \centering
                \includegraphics[width=0.9\textwidth]{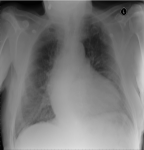}
                \caption{}
                \label{fig:acuracy}
        \end{subfigure}%
        \begin{subfigure}[t]{0.12\textwidth}
        \centering
                \includegraphics[width=0.9\textwidth]{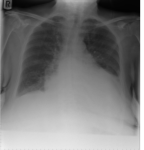}
                \caption{}
                \label{fig:loss}
        \end{subfigure}%
        \begin{subfigure}[t]{0.12\textwidth}
        \centering
                \includegraphics[width=0.9\textwidth]{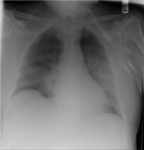}
                \caption{}
                \label{fig:acuracy}
        \end{subfigure}%
        
        \begin{subfigure}[t]{0.12\textwidth}
        \centering
                \includegraphics[width=0.9\textwidth]{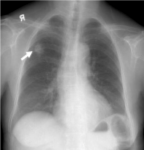}
                \caption{}
                \label{fig:acuracy}
        \end{subfigure}%
        \begin{subfigure}[t]{0.12\textwidth}
        \centering
                \includegraphics[width=0.9\textwidth]{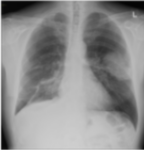}
                \caption{}
                \label{fig:acuracy}
        \end{subfigure}%
        \begin{subfigure}[t]{0.12\textwidth}
        \centering
                \includegraphics[width=0.9\textwidth]{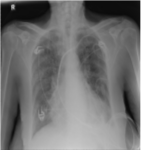}
                \caption{}
                \label{fig:loss}
        \end{subfigure}%
        \begin{subfigure}[t]{0.12\textwidth}
        \centering
                \includegraphics[width=0.9\textwidth]{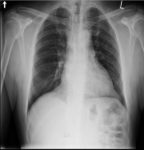}
                \caption{}
                \label{fig:acuracy}
        \end{subfigure}%
        
        \begin{subfigure}[t]{0.12\textwidth}
        \centering
                \includegraphics[width=0.9\textwidth]{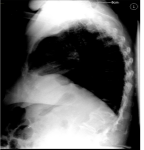}
                \caption{}
                \label{fig:loss}
        \end{subfigure}%
        \begin{subfigure}[t]{0.12\textwidth}
        \centering
                \includegraphics[width=0.9\textwidth]{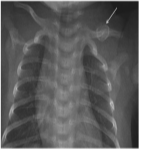}
                \caption{}
                \label{fig:acuracy}
        \end{subfigure}%
        \begin{subfigure}[t]{0.12\textwidth}
        \centering
                \includegraphics[width=0.9\textwidth]{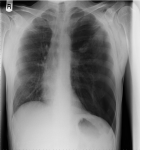}
                \caption{}
                \label{fig:acuracy}
        \end{subfigure}%

        \caption{Example chest radiographs for each categories. a)~Normal; b)~Cardiomegaly; c)~Consolidation; d)~Pulmonary Edema; e)~Lung Nodules; f)~Lung Mass; g)~Pleural Effusion; h)~Widened Mediastinum; i)~Vertebral Fractures; j)~Clavicular Fracture; k)~Pneumothorax.}
        \label{fig:chest_images} 
         \vspace{-4mm}

\end{figure}

\begin{table*}[]
\centering
\caption{A general consensus for the breast density descriptors for five categories of breast density.}
\captionsetup{font=small}

\label{T:general_consensus_mamo}
\begin{adjustbox}{width=1\textwidth,center=\textwidth}
\begin{tabular}{|l|c|c|c|l|}
\hline
\multicolumn{1}{|c|}{Category}& Label  & Count & Breast Density& \multicolumn{1}{c|}{Example Breast Density Descriptors}                           \\ \hline
\hline
Almost entirely fatty&0 &342&   $<25\%$     & fat; fatty; includes variations of mainly fatty; predominantly fatty;  predominantly fat \\ \hline
Scattered areas of fibroglandular density &1 &1546 &  $25-50\%$      & scattered                                \\ \hline
Heterogeneously dense &2 &  1510&    $51-75\%$     & heterogenous; heterogeneously                                     \\ \hline
Extremely dense &3 &  436 &  $>75\%$  & dense; extremely dense; very dense                                                 \\ \hline
Ambiguous &4 & 246 & - & mild dense; mildly dense; overlap in describing categories with label 1 and 2                            \\ \hline
\end{tabular}
\end{adjustbox}
\label{T:breast_consensus}
\end{table*}

\begin{table*}[]
\centering
\caption{A sample of radiology reports with associated categories for mammogram reports dataset (MRD).}
\captionsetup{font=small}

\label{T:breast_sample}
\begin{adjustbox}{width=1\textwidth,center=\textwidth}
\begin{tabular}{|l|l|}
\hline
\multicolumn{1}{|c|}{Category}& \multicolumn{1}{|c|}{Example from MRD}                                                                                                                                                                                                 \\ \hline
Almost entirely fatty & \begin{tabular}[t]{@{}l@{}}The breasts are almost entirely fatty bilaterally. No suspicious calcification, dominant mass or architectural distortion. No interval change.\end{tabular}   \\ \hline

Scattered areas of fibroglandular density &The breasts show scattered fibroglandular densities. There are no dominant nodules or suspicious calcification seen in either breast. \\ \hline

Heterogeneously dense & The breast parenchyma is heterogeneously dense. No suspicious masses, calcification or architectural distortion seen on either side.\\ \hline

Extremely dense  & The breast parenchyma is extremely dense. No suspicious masses, calcifications or architectural distortion seen on either side.\\ \hline

\end{tabular}
\end{adjustbox}
\end{table*}

\begin{table*}[]
\centering
\caption{A general consensus for the chest radiographs descriptors for 11 categories.}
\captionsetup{font=small}

\label{T:general_consensus_mamo}
\begin{adjustbox}{width=1\textwidth,center=\textwidth}
\begin{tabular}{|l|c|c|l|}
\hline
\multicolumn{1}{|c|}{Category}& Label  & Count &  \multicolumn{1}{c|}{Example Chest Radiograph Descriptors}                           \\ \hline
\hline
Normal &0 &116&   No acute findings; unremarkable study; the cardiopericardial silhouette and hilar anatomy is within normal limits \\ \hline
Cardiomegaly &1 &106      & Mild cardiomegaly/pericardial effusion; enlarged cardiac silhouette; cardiopericardial silhouette is enlarged                                \\ \hline
Consolidation &2 &  81& Resolving consolidation in the lower  lungs bilaterally, with overall improved aeration in the lower lungs                                  \\ \hline
Pulmonary Edema&3 &  104 &  Mild to moderate interstitial pulmonary edema; airspace edema; right pulmonary edema                                                   \\ \hline
Lung Nodules&4 & 118 & Calcified granulomas present; nodules have developed; multiple faint bilateral pulmonary nodules                              \\ \hline
Lung Mass&5 & 89 & Bilateral pulmonary masses; mass in the right upper lobe; middle mediastinal mass                     \\ \hline
Pleural Effusion&6 & 144 &  Persistent bilateral pleural effusions; bilateral pleural effusions; persistent loculated right pleural effusion                             \\ \hline
Widened Mediastinum&7 & 50 & Widening of the superior mediastinum; mediastinum appears widened; mediastinum is slightly widened                         \\ \hline
Vertebral Fractures&8 & 67 &  Vertebral fracture; several old vertebral compression injuries; thoracolumbar vertebral compression injuries                             \\ \hline
Clavicular Fracture&9 & 73 & Fracture deformity of the right lateral  clavicle; right lateral clavicle fracture                       \\ \hline
Pneumothorax&10 & 82 & Partial right upper lobe collapse; chronic collapse of the right middle lobe; right apical pneumothorax                            \\ \hline

\end{tabular}
\end{adjustbox}
\label{T:chest_consencus}
\end{table*}

\begin{table*}[]
\centering
\caption{A sample of radiology reports with associated categories for chest radiographs dataset (CRRD).}
\captionsetup{font=small}

\label{T:chest_sample}
\begin{adjustbox}{width=1\textwidth,center=\textwidth}
\begin{tabular}{|l|l|}
\hline
\multicolumn{1}{|c|}{Category}&\multicolumn{1}{|c|}{Example CRRD}           \\ \hline

Normal &\begin{tabular}[t]{@{}l@{}}Cardiomediastinal contours are within normal limits. The hila are unremarkable. The lungs are clear.   \end{tabular}             \\ \hline
Cardiomegaly & \begin{tabular}[t]{@{}l@{}} The cardiopericardial silhouette is enlarged with LV prominence. Aortic  valve prosthesis in situ. There is unfolding of the aorta with  calcification.  \end{tabular}             \\ \hline
Consolidation &\begin{tabular}[t]{@{}l@{}}The lungs remain overinflated and show mild chronic parenchymal changes. Inhomogeneous airspace consolidation has developed \\in the basal segments of the left lower lobe. 
 \end{tabular}      \\ \hline
Pulmonary Edema&\begin{tabular}[t]{@{}l@{}}Significant fluid has developed in the right minor fissure. Pulmonary  venous markings are severely elevated.\end{tabular}\\ \hline 
 Lung Nodules &\begin{tabular}[t]{@{}l@{}}The left pulmonary nodular density superimposed on the posterior seventh rib is smaller. \\There is mild stable bilateral upper lobe pleuropulmonary scarring and a stable right upper lobe nodule.                                                          \end{tabular}       \\ \hline
 Lung Mass&\begin{tabular}[t]{@{}l@{}}Left upper lobe perihilar mass. Prominence of the left hila  as well as the left superior mediastinal margins. \end{tabular}\\ \hline
 Pleural Effusion&\begin{tabular}[t]{@{}l@{}} Midline sternotomy wires appear stable in position. Small bilateral  pleural effusions unchanged from previous exam. \end{tabular}\\ \hline
 Widened Mediastinum&\begin{tabular}[t]{@{}l@{}}Superior mediastinal  widening and vascular engorgement relate to patient positioning. The lungs are clear.\end{tabular}\\ \hline
 Vertebral Fractures &\begin{tabular}[t]{@{}l@{}}The cardiac silhouette is normal in size and shape with aortic unfolding. The lungs are overinflated but clear. \\The pleural spaces, mediastinum and diaphragm appear normal. 
\end{tabular}\\ \hline
  Clavicular Fracture&\begin{tabular}[t]{@{}l@{}}Degenerative changes in the spine and an old right mid clavicular fracture.\end{tabular}\\ \hline
 
 Pneumothorax &\begin{tabular}[t]{@{}l@{}}Right-sided pleural drain in situ. There is very subtle residual right  pneumothorax with lateral pleural edge displaced. \end{tabular}\\ \hline
\end{tabular}
\end{adjustbox}
\end{table*}

\section{The Datasets}
\label{sec:data}
Our institutional review board approved this single-center retrospective study with a waiver for informed consent. The Research Ethics Boards (REB) number is 17-167. A search of our Radiology Information System (RIS) (Syngo; Siemens Medical Solutions USA Inc, Malvern, PA) was preformed for mammography and chest radiograph reports using Montage Search and Analytics (Montage Healthcare Solutions, Philadelphia, PA). This search identified 4,080 mammography reports and 1,030 chest radiographic reports (see Table~\ref{T:stats}). These reports were exported from the RIS and removed of any patient identifying information.

\subsection{Mammogram Reports Dataset}



The mammography reports dataset (MRD) contained 4,080 reports. Breast density was classified in each report according to the American College of Radiology Breast Imaging Reporting and Data System (BI-RADS) classification system \cite{taco1998american} 
(Figure~\ref{fig:breast_images}). Breast density reflects the relative composition of fat and fibroglandular tissue. Almost entirely fatty refers to breasts with less than $25\%$ areas of fibroglandular tissue. Scattered ares of fibroglandular density is used for breasts composed of $25-50\%$ fibroglandular tissue. Heterogeneously dense breasts are composed of $50-75\%$ fibroglandular tissue and the extremely dense breasts have greater than $75\%$ of fibroglandular tissue density. Descriptive statistics of this dataset are presented in Table~\ref{T:breast_consensus} and sample reports are presented in Table~\ref{T:breast_sample}.

\subsection{Chest Radiographs Reports Dataset}
The chest radiograph dataset (CRRD) consisted of 1,030 reports. These reports were classified into 11 categories which includes normal examinations and ten pathologic states (Figure~\ref{fig:chest_images}). Descriptive statistics of this dataset are presented in Table~\ref{T:chest_consencus} and sample reports are presented in Table~\ref{T:chest_sample}. This dataset has more categories, fewer reports per category, and is less imbalanced when compared with the MRD. 

%

\begin{figure*}[!htbp]
\centering
        \begin{subfigure}[t]{0.33\textwidth}
        \centering
                \includegraphics[width=1.05\textwidth]{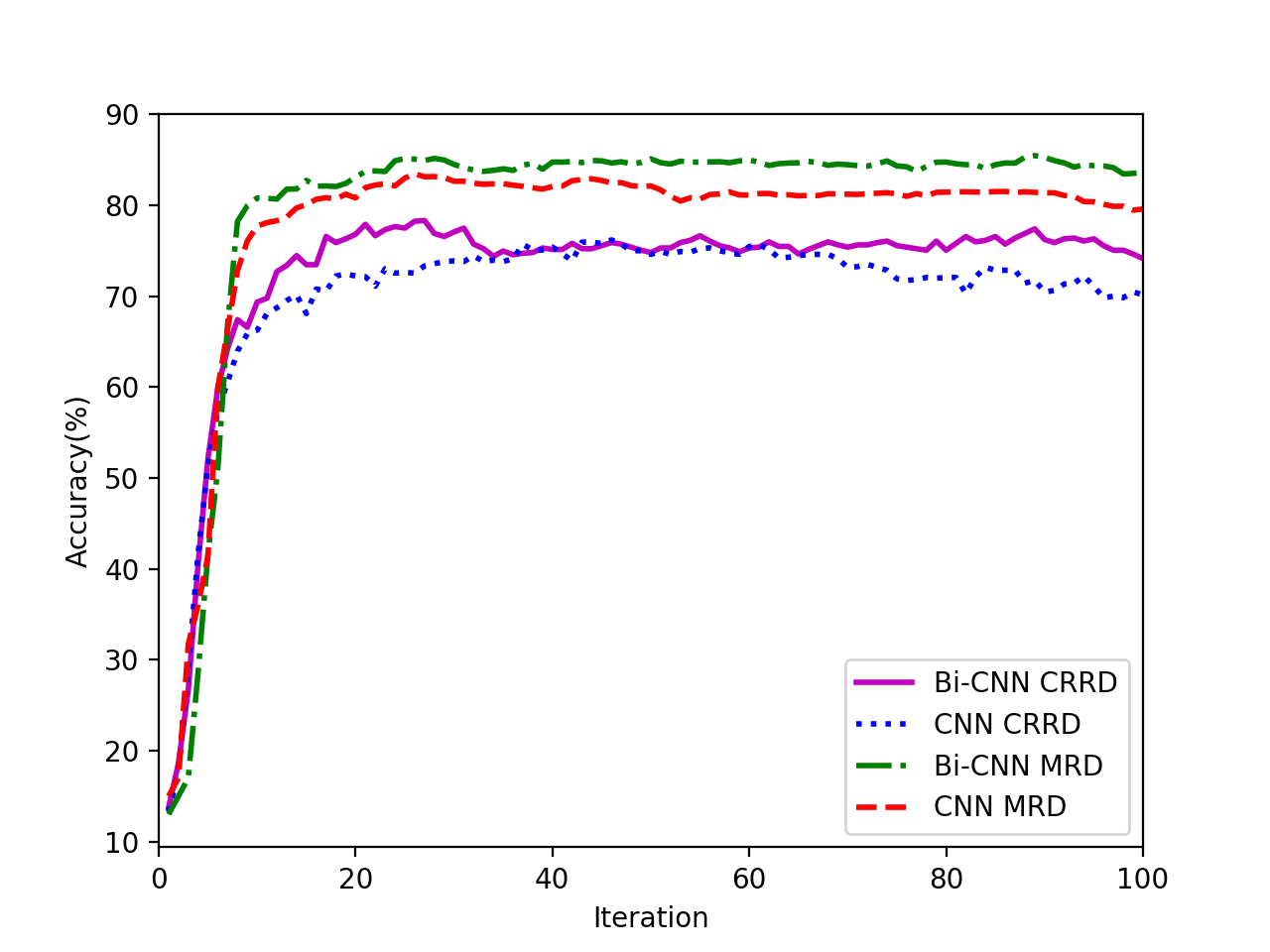}
                \caption{$\lambda=0.1$.}
        \end{subfigure}%
        \begin{subfigure}[t]{0.33\textwidth}
        \centering
                \includegraphics[width=1.05\textwidth]{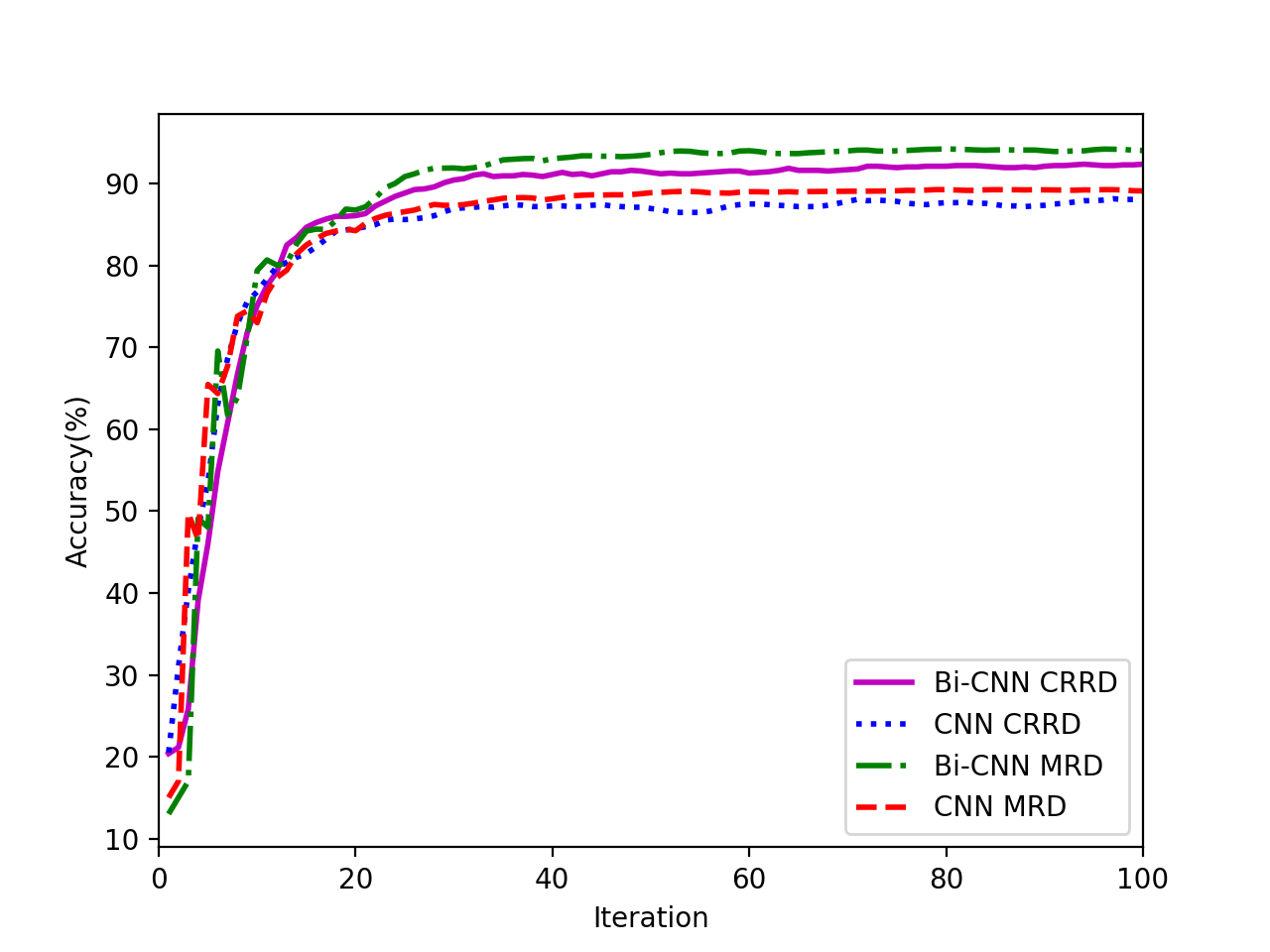}
                \caption{$\lambda=0.01$.}
        \end{subfigure}%
        \begin{subfigure}[t]{0.33\textwidth}
        \centering
                \includegraphics[width=1.05\textwidth]{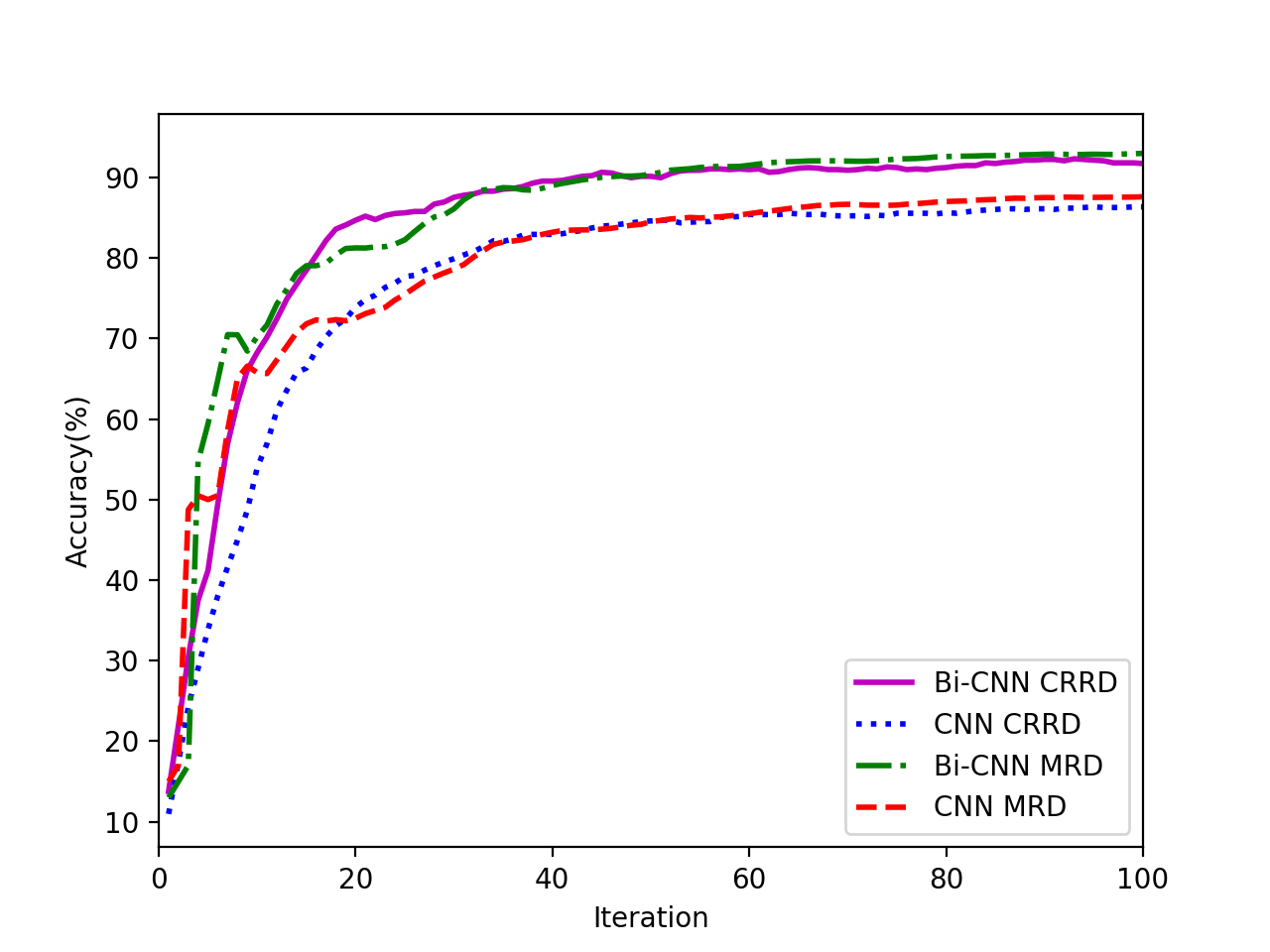}
                \caption{$\lambda=0.001$.}
        \end{subfigure}%

\caption{Accuracy (in $\%$) and convergence behaviour of CNN and Bi-CNN models for the MRD and CRRD for different values of learning rate $\lambda$. The x-axis represents the training iteration. The plots are from cross-validation over 30 independent experiments.}
\label{fig:acc}
 \vspace{-4mm}
\end{figure*}

\section{Experiments and Results Analysis}
\label{sec:results}

In this section, we compare performance of the proposed Bi-CNN model with the random forest (RF), SVM, and CNN models. The experiments were conducted on both the mammogram and chest radiograph datasets. 

The CNN and Bi-CNN models are implemented in TensorFlow \cite{abadi2016tensorflow} and the RF and SVM models are implemented using the classifiers in scikit-learn \cite{pedregosa2011scikit}. The experiments are conducted on a DevBox with an Intel Core i7-5930K 6 Core 3.5GHz desktop processor, 64 GB DDR4 RAM, and two TITAN X GPUs with 12GB of memory per GPU.

\subsection{Parameters Setting}
Unless stated, the CNN and Bi-CNN models are trained with a mini-batch size of 64, drop-out probability of 0.5, filter sizes \{3, 4, 5\} and 120 feature maps per filter size. The number of training iterations is set to 50 and the initial learning rate for Adam optimizer is set to 0.001 \cite{kingma2014adam}. 
An exponential decay adaptive learning rate is applied. The weights at output layers are initialized 
using the Xavier method \cite{glorot2010understanding}, the weights in the convolutional layer are selected based on normal distribution with standard deviation of 0.1, and biases are set to 0.1. The activation function before the max-pooling layer is ReLU \cite{nair2010rectified}. The $L_{2}$ regularization is set to $1.0 \times 10^{-4}$ and early-stopping is applied.

For all the experiments, $70\%$, $15\%$, and $15\%$ of data is allocated for training, validation, and test, respectively. The data is shuffled before splitting. In each experiment, the model is cross-validated over 30 independent experiments and the results after statistical testing are reported. 

\subsection{Performance Evaluation and Analysis}
We experimented with various configurations of the RF \cite{breiman2001random}, SVM \cite{hearst1998support}, and CNN \cite{kim2014convolutional} models.

\begin{itemize}  
\item \textbf{RF}: Instead of using a bag-of-the-words technique, which uses the frequency of the words in a query as the features, a $n$-gram model with a lower boundary of 1 and upper boundary $n\in\{1,2,3\}$ performs the feature extraction of reports. We use an implementation of RF \cite{breiman2001random} which combines classifiers by averaging their probabilistic predictions. The number of estimators is set to 10. 

\item \textbf{SVM}: Similar to RF, the SVM model uses $n$-gram feature extraction. The SVM model is deployed for two different kernels, a ``sigmoid" and a ``polynomial" with degree three.

\item \textbf{CNN}: The CNN implementation proposed in \cite{kim2014convolutional} for text classification and later used in \cite{shin2017classification} for the classification of radiology head CT reports is used. This model has a single input channel and the order of input words is similar to the order in the report. The model has a convolution layer followed by a max-pooling layer and a ``softmax" classifier. The studies are for kernel sizes $k\in\{1,2,3\}$.

\item \textbf{Bi-CNN}: The proposed Bi-CNN with two input channels. The settings are similar to CNN.
\end{itemize}

The experiments are for breast density classification and chest radiograph classifications based on chest pathology using the MRD and CRRD, respectively. 
The results of performance comparisons between the RF, SVM, CNN, and Bi-CNN models are presented in Table~\ref{T:performance}. As $n$ increases in $n$-gram for RF, it considers the dependency between a greater number of words (i.e. $n$). The RF with 3-gram model has better performance than the 2-gram and 1-gram RF models. The SVM with polynomial of degree three (i.e., ``poly") and ``sigmoid" kernels almost have the same performance. The number of kernels in the CNN models have a minor impact on accuracy. However, the CNN models with 3-kernel have better performance than models with 1-kernel and 2-kernel. The proposed Bi-CNN with 3-kernel has the best performance compared to the other models.

\begin{table}[]
\centering
\caption{Classification accuracy (in $\%$) for MRD and CRRD using RF, SVM, CNN, Bi-CNN. The best results are in boldface.}
\captionsetup{font=small}
\label{T:performance}
\begin{tabular}{|l|c|c|}
\hline
\multicolumn{1}{|c|}{Model} & MRD   & CRRD  \\ \hline
RF (1-gram)                 & 69.73 & 67.32 \\ \hline
RF (2-gram)                 & 73.76 & 69.27 \\ \hline
RF (3-gram)                 & 80.53 & 77.59 \\ \hline
SVM (sigmoid)               & 82.74 & 80.24 \\ \hline
SVM (poly)                  & 82.97 & 81.73 \\ \hline
CNN (1-kernel)              & 85.72 & 85.27 \\ \hline
CNN (2-kernel)              & 86.52 & 86.01 \\ \hline
CNN (3-kernel)              & 86.91 & 86.05 \\ \hline
Bi-CNN (1-kernel)           & 89.60 & 89.16 \\ \hline
Bi-CNN (2-kernel)           & 90.88 & 89.91 \\ \hline
Bi-CNN (3-kernel)           & \textbf{92.94} & \textbf{91.34} \\ \hline
\end{tabular}
 \vspace{-1mm}
\end{table}

\begin{table}[]
\centering
\caption{Accuracy (Acc. in $\%$) and convergence iteration (C) of CNN and Bi-CNN models for learning rate
$\lambda$ on MRD and CRRD datasets. Best results are in boldface.}
\begin{adjustbox}{width=0.48\textwidth}

\begin{tabular}{|c|c|c|c|c|c|c|c|c|}
\hline
\multirow{2}{*}{Model} &\multicolumn{4}{c|}{CNN} &\multicolumn{4}{c|}{Bi-CNN} \\ \cline{2-9}
   &\multicolumn{2}{c|}{MRD}&\multicolumn{2}{c|}{CRRD} &\multicolumn{2}{c|}{MRD}&\multicolumn{2}{c|}{CRRD} \\ \hline

$\lambda$ & Acc.  & C & Acc. & C & Acc.  & C & Acc. & C \\ \hline
\hline
 0.1 &  82.54 & 36 &76.42 & 43 & \textbf{90.03} &\textbf{23}&\textbf{84.04} & \textbf{26} \\ \hline
 0.01 &  83.27 & 100 & 82.78 & 100 & \textbf{89.98 }& \textbf{100} & \textbf{87.85 }& \textbf{100}     \\ \hline
0.001 &   86.91 & 100 & 86.05 & 100 & \textbf{92.94} & \textbf{100} & \textbf{91.34} & \textbf{100}      \\ \hline

\end{tabular}
\end{adjustbox}
\label{T:lr_analysis}
 \vspace{-3mm}
\end{table}

The convergence behaviour of CNN and Bi-CNN over training iterations for the validation dataset is presented in Figure~\ref{fig:acc}. For a large learning rate value (i.e., $\lambda=0.1$), the models converge rapidly to a local solution. However, for smaller learning rates, the models converge more slowly but with greater stability towards a better solution. As the results in Table~\ref{T:lr_analysis} show, the CNN model converges with a lower accuracy for three different learning rates $\lambda\in\{0.1,0.01,0.001\}$. For $\lambda=0.001$, the best performance is achieved by Bi-CNN for both MRD and CRRD. One of the advantages of Bi-CNN is adding diversity into the model through the integration of two different representations of feature vectors into the model. This diversity helps the CNN avoid convergence to local solutions with low accuracy.

\section{Conclusion}
\label{sec:conclusion}
Radiology reports contain a radiologist's interpretation of a medical imaging examination. Vast numbers of such reports are digitally stored in health care facilities. Text mining and knowledge extraction from such data may potentially be used for double checking radiologist interpretations as part of an audit system. Furthermore, these methods can be used to audit utilization of limited health care resources and to enhance the study of patient treatment plans over time.

In this paper, we have proposed a Bi-CNN for the interpretation of radiology reports. We have collected and anonymized two datasets for our experiments: a mammogram reports dataset and a chest radiograph reports dataset. The Bi-CNN has two input channels where one channel represents the features of the report (including zero-padding) and the other channel represents the reverse non-padded features of the report. The combination of forward and reverse order of feature vectors represents more information about dependencies between feature vectors. Our comparative studies demonstrate that the Bi-CNN model outperforms the CNN, RF, and SVM methods.


%

\section*{Acknowledgement}
The authors thank the support of NVIDIA Corporation with the
donation of the Titan X GPUs used for this project.

\bibliographystyle{IEEEtran}
\bibliography{CTLIEEEtrans,mybibfile}

%

%
%
%




\end{document}